# UAVs for Industries and Supply Chain Management

**Shrutarv Awasthi**
Scientific Assistant
TU Dortmund
Chair of Material Handling and Warehousing

**Nils Gramse**
Scientific Assistant
TU Dortmund
Chair of Material Handling and Warehousing

**Dr. Christopher Reining**
Chief Scientist
TU Dortmund
Chair of Material Handling and Warehousing

**Dr. Moritz Roidl**
Chief Engineer
TU Dortmund
Chair of Material Handling and Warehousing

*This work aims at showing that it is feasible and safe to use a swarm of Unmanned Aerial Vehicles (UAVs) indoors alongside humans. UAVs are increasingly being integrated under the Industry 4.0 framework. UAV swarms are primarily deployed outdoors in civil and military applications, but the opportunities for using them in manufacturing and supply chain management are immense. There is extensive research on UAV technology, e.g., localization, control, and computer vision, but less research on the practical application of UAVs in industry. UAV technology could improve data collection and monitoring, enhance decision-making in an Internet of Things framework and automate time-consuming and redundant tasks in the industry. However, there is a gap between the technological developments of UAVs and their integration into the supply chain. Therefore, this work focuses on automating the task of transporting packages utilizing a swarm of small UAVs operating alongside humans. MoCap system, ROS, and unity are used for localization, inter-process communication and visualization. Multiple experiments are performed with the UAVs in wander and swarm mode in a warehouse like environment.*

***Keywords:** (*UAV swarm, Warehousing, Transportation, Logistics, supply chain, Industry 4.0*)*

## 1. INTRODUCTION

Logistics is a fundamental part of an industry as it ensures the link between suppliers, manufacturers, and customers [1]. In industry 4.0, the logistic sector is being digitized and automated to make the supply chain more efficient. In the EU, warehousing and storage represent up to 15% of the current costs in logistics [2]. Audi has tested the use of Unmanned Aerial Vehicles (UAVs) for automated transport of parts in factory halls [3]. It is speculated that the UAVs will be rapidly integrated into the logistics and supply chain, and this market will reach 29$ billion by 2027 with a compound yearly growth rate of nearly 20% [4].Thus, there is a growing need to optimize and automate the operations in warehouses and industrial facilities using UAVs. In addition, there is a scarcity of research on using UAV swarms indoors in confined and crowded industrial environments alongside humans. Modern order picking warehouse layouts mostly consists of multiple tall racks with narrow aisles in between [5]. This makes it challenging and time-consuming for humans to manage the inventory. Small and lightweight UAVs, can traverse in a cluttered and constrained indoor environment, thus having significant benefits for optimizing the supply chain [6].

The supply chain has gradually become the central organizing unit in industries [7]. As a result, enterprises are constantly confronted with the need to efficiently manage the ever-increasing supply chain activities. UAVs are capable of revolutionizing and transforming the supply chain. They can speed up and streamline processes, lower costs, and reduce environmental impact. UAVs offer promising and cost-efficient applications such as inspection of hard-to-reach or hazardous areas, detection of gas leaks in large plants, inventory management, stock taking and transporting goods from shelves to picking stations [8] [9] [10].

UAVs also fit perfectly into the Industry 4.0 setting, where all devices are networked and can share their data [9]. UAVs can operate in the unused space above the production systems and thus extend the flow of materials to the third dimension [8]. Most research on UAVs in logistics focuses on using a single UAV indoors or outdoors [5] [10]. Many works use UAV swarms but focus on outdoor logistics and transportation [11] [12]. There is little research using a swarm of UAVs indoors, focusing on transporting goods in a warehouse or an industrial setting [11] [12].

In this work, we deploy a swarm of UAVs alongside humans to automate the task of picking up multiple packages from various locations and delivering them to the specified destination. The task is performed in a warehouse-like environment using Crazyflie UAVs. The UAVs are small and lightweight and thus incapable of carrying an additional load; therefore, we demonstrate our method using virtual packages.

## 2. RELATED WORKS

Over the past years, the application of UAVs in the transportation and indoor logistics sector has become the focus of various researches. Villa et al. review the load transportation using UAVs [12].

Lieret et al. present an automated transportation of small packages with UAVs inside a warehouse and manufacturing areas. The UAV first navigates to a transfer area where the package is automatically loaded onto it. Subsequently, the UAV transports the package to the destination and releases it without any human intervention. The authors implement the task of transportation on a single UAV [8].

Alshanbari et al. equip UAVs with artificial intelligence (AI) capabilities to achieve autonomous transportation of packages. The camera on the drone identifies the target location and releases the package once it reaches the target. The authors tested their approach on a UAV in an indoor and outdoor environment [13]. Boshoff et al. presented a workflow for automated transportation of terminals and jumpers by UAVs. The authors considered the terminal strip assembly line to examine the need for automated picking of terminals and jumpers by UAVs. They concluded that integrating an automated UAV-based supply into the assembly line might transform the assembly line to a more flexible structure and reduce production effort [14].

Marquez-Vega Et al. implement a target zone search without and with the presence of obstacles in an arena using UAV swarms of varying sizes [15]. Kim et al. presents an efficient exploration strategy to explore unknown environments using a swarm of UAVs [16]. Cristiani et al. use a swarm of small Parrot UAVs to automatize inventory management. The authors propose a generic architecture for UAV swarms sweeping a warehouse and identifying the packages at each shelf unit employing QR code technology. They integrate a ground charging station to enable autonomous charging of UAVs. Based on simulations, an analysis is done in terms of package recognition accuracy and inventory completion time [5].

## 3. ARCHITECTURE

The UAVs used in this work use the Crazyflie 2.0 as a base platform [17] . The UAVs are modified to increase their speed and payload. A single UAV has a size of 103x103x29 mm and a total weight of 37.85 g. Figure 1 shows the UAV used in this work. The Robot Operating System (ROS), a middleware, is responsible for handling the inter-process communication between the MQTT server and UAVs [18]. It enables one to abstract the hardware (UAVs) from the software. ROS is mainly developed using two languages: C++ and Python. The objects in the test area are marked with a unique combination of markers, enabling the MoCap system to identify and provide the absolute position and attitude of the objects in a three-dimensional space. The MoCap system consists of 52 infrared cameras manufactured by Vicon. The MoCap sends the pose and attitude of every marked object to the MQTT server. Unity gets the poses and attitudes from MQTT, enabling all marked objects to be mirrored into the Unity 3D simulation environment.

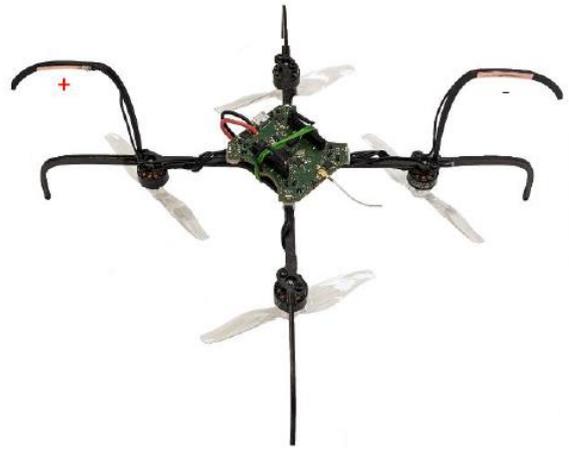

**Figure 1. Crazyflie UAV**

The Unity subsystem implements the swarm behavior as a state machine. Algorithms for wandering, alignment, cohesion, separation, pursuit, and fencing are implemented in unity to ensure a safe and collision-free trajectory for the UAVs. Unity sends simple motion set-points to the UAVs via MQTT and receives the updated positions of the UAVs via the MoCap connection. The low-level control logic is performed onboard the UAVs. Moreover, the UAV trajectory is computed onboard using the set-point and the state estimates.

The simulation subsystem comprises a laser projection system consisting of eight Kvant Clubmax FB4 laser projectors. The laser system generates both static and dynamic projections of virtual objects from the simulation. Therefore, it is possible to visualize the transportation task explained in this work [19]. Visualization is done for demonstration purposes and a better understanding of the complex behavior of the algorithms. Virtual objects such as packages, package delivery destinations and obstacles are mirrored on the area floor via the laser projection system.

Communication between the ROS server and the drone takes place via the 2.4 GHz ISM band. The CRTP (Crazyflie Real-Time Protocol) sends data packets without much overhead, thus minimizing the latency. The transmission speed is configured to 250 Kbit/s to prevent interference caused by the metal walls of the warehouse and enable the transmitted signals to be decoded reliably by the UAVs. The architecture shown in Figure 2 creates an environment for simulating real warehouse scenarios. UAVs and their trajectories can be simulated at an accelerated time in Unity. Real-time data of the UAV swarm and objects in the test area are utilized to simulate a warehouse scenario. Thus, creating a digital twin.

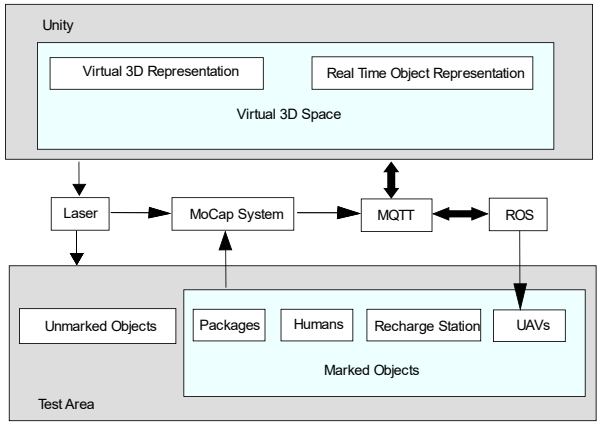

**Figure 2. Architecture used for automating and visualizing the task of transporting multiple packages from source to destination using a UAV swarm.**

## 4. METHOD

In this work, the task of transporting multiple virtual packages from various pick-up locations and delivering them to a specified destination is performed using a UAV swarm working alongside humans in a warehouse. The entire process is also visualized in real time using a Unity and Laser subsystems.

There are two modes for the UAVs to operate: a) Wander and b) Swarm. In the swarm mode, the UAVs arrange themselves in either a single big group or multiple small-sized groups and remain in that formation. Every group is an independent swarm. A swarm can break into smaller groups due to an obstacle. UAVs in a swarm are controlled using the swarm algorithms (alignment, cohesion, separation) [20]. In Wander mode, UAVs fly independently in the test area.

The swarm behavior, established by Reynolds, is implemented using three basic rules: cohesion rule, alignment rule, and separation rule [21] shown in Figure 3. The black triangle represents the active UAV, and the remaining triangles represent other UAVs. The black arrow indicates the resulting vector adapted to satisfy the requirements of a rule. The white circle visualizes the effective range of local perception of the active UAV. The black circle represents the center of all UAVs located in the local perception.

The cohesion rule aims to move the active UAV towards the swarm's center. Thus, the UAV tries to point its velocity vector to the center of its local perception. The alignment rule ensures that all UAVs within the range of local perception align themselves to move in the same direction. Finally, the separation rule states that the active UAV maintains a minimum distance to all its neighboring UAVs [21]. Weighting the individual rules influences the appearance of the swarm. For example, if collisions must not occur under any circumstances, the separation rule is given a high weighting. After assigning appropriate weights, the individual velocity vectors are summed up to a control vector, which determines the final movement of an agent, as can be seen in Figure 3. The separation rule is active in the swarm and the wander mode, whereas the cohesion and alignment rule is active only in the swarm mode. Thus, UAVs are able to avoid obstacles and inter UAV collisions.

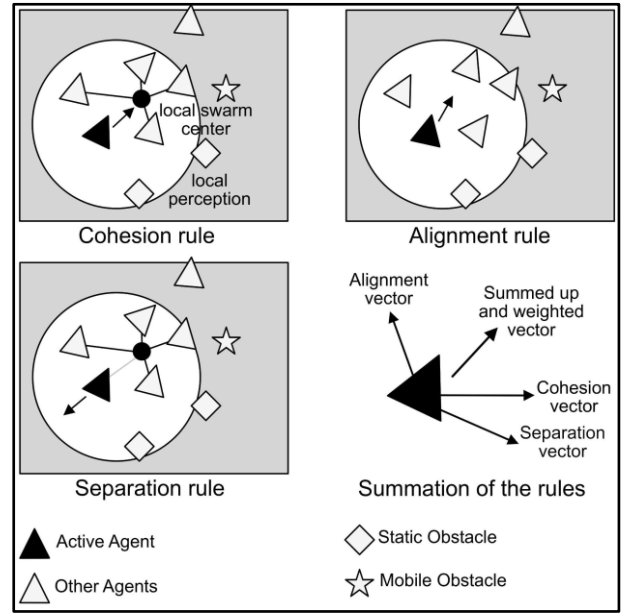

**Figure 3. Schematic representation of the three basic rules for simulated swarm behavior** [21]**. Mobile obstacles are people moving with headgear, and static obstacles could be the charging station, walls, or shelves.**

A person (marked) in the test area is responsible for controlling the UAV swarm. Initially, on the person's command, the UAVs start from a fixed location in the test area and operate in the wander mode. The person can switch the mode of the UAVs between the swarm and the wander mode. The UAVs can detect a package irrespective of their flying modes. When a marked disk touches the floor of the test area, a virtual package is created. Multiple virtual packages can be created using the disk. As soon as a package is created, the UAV closest to it flies to the package, hovers a few cm above it for a few seconds and flies to the destination while maintaining the same height. After virtually transporting the package to the destination, the UAV flies higher and starts to operate in its previous mode. The hovering time is the time the drone would require to pick up the package. Laser projections on the floor help to visualize the entire task.

To ensure safety, a UAV automatically lands if its battery is low or in case of a communication failure. Communication failure means a UAV does not receive a data packet from the ROS server for 3 seconds. If a UAV fails or lands during operation, the remaining UAVs in the swarm continue the task without interruption. Thus, the architecture is flexible. Finally, when the person gives the land command, the UAVs land at the location from where they had started initially.

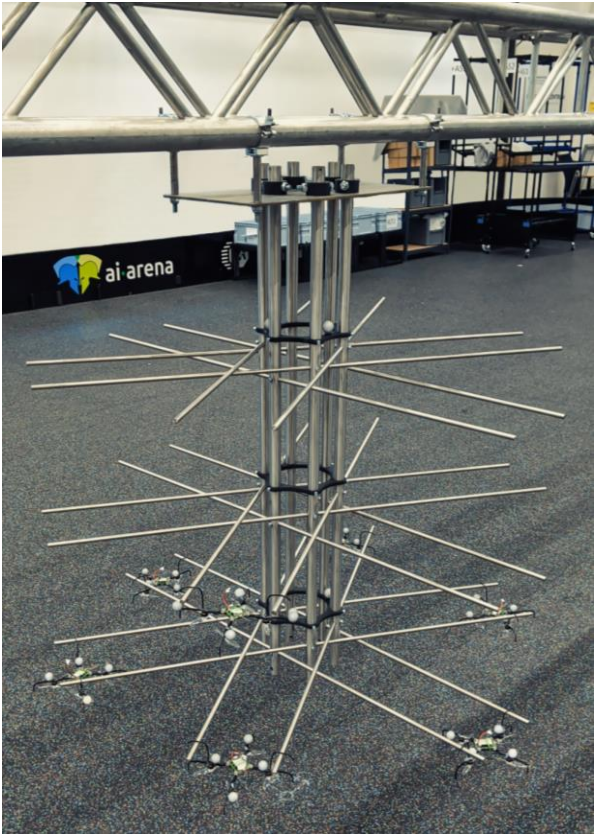

**Figure 4. Recharge station.**

Multiple experiments were performed with a swarm of 16 UAVs around humans in the testing area and not once did the UAVs collide with the humans. Moreover, in each experiment, the UAVs completed the transportation task. Thus, the UAV swarm can be used in a warehouse for transporting packages from pick-up bins to dedicated shelves. It is also safe for humans to work alongside UAVs. A UAV recharge station, shown in Figure 4, has been built to enable UAVs to recharge autonomously when low on battery but has not yet been integrated into the unity system. The main idea behind creating the recharge station is to enable the 24/7 operation of the swarm. The battery of every UAV is continuously monitored, and when a UAV is low on battery, that UAV automatically flies to the station for charging. Another fully charged UAV from the station replaces it. The UAVs have hooks to hang from the charging station, as seen in Figure 1. The polarity of the hooks is pre-defined, and copper strips are attached to them to charge the battery. The stations can be hung from the ceiling or placed securely in the warehouse.

## 5. APPLICATION

A possible logistics scenario for a UAV swarm, especially with the ability to recharge independently, is their use as a mobile surveillance system. The UAV swarm can be used in warehouses, for example, to protect the stored goods from theft. The drones can then perch like birds on the existing trusses, high racks and steel beams and observe their surroundings from the high positions. They can gather a more significant amount of data instead of low power stationary IoT devices equipped with a battery.

Another possible application is in a smart factory, where humans signal the UAV swarm to get a tool. The UAV picks the tool from the toolbox and delivers it to the person. Thus, automating the task of picking up and placing objects using a UAV in an industry will save time and avoid congestion among workers.

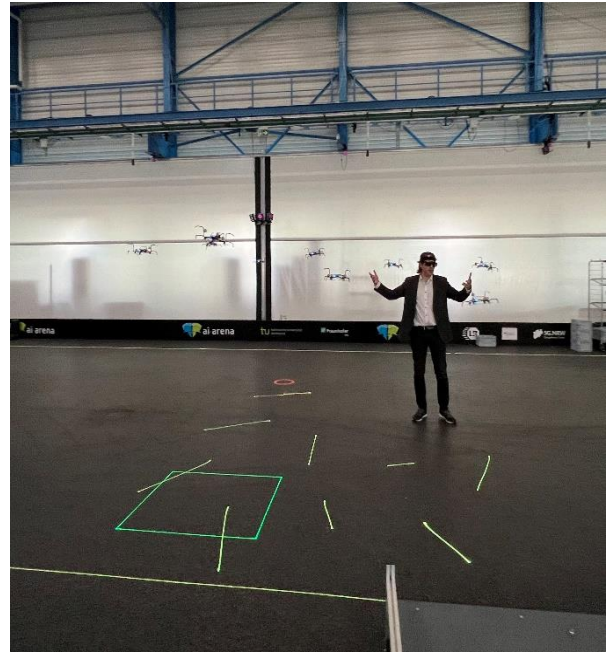

**Figure 5. Experimental setup showing UAVs operating in swarm mode in the test area.**

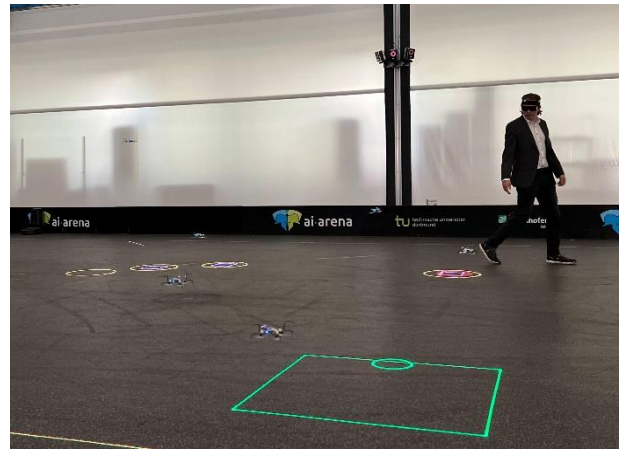

**Figure 6. Experimental setup showing UAVs in wander mode performing the transportation task alongside humans in the test area.**

## 6. CONCLUSION

In this work, a swarm of UAVs is deployed alongside humans to automate the task of picking up multiple virtual packages from various locations and delivering them to the specified destination. The task is performed in a warehouse-like environment using Crazyflie UAVs. Furthermore, a digital twin is created using Unity and Laser subsystems, enabling one to simulate a warehouse-like environment. Experiments show that it is feasible and safe to use a UAV swarm indoors alongside humans. Moreover, a UAV swarm has tremendous potential to automate various industrial tasks and optimize the supply chain safely. Thus, in the future, UAVs will play a significant role in revolutionizing industries' operations.

In the future, we plan to use bigger UAVs to carry actual packages and integrate the recharging station with the current scenario. Furthermore, further research on techniques for indoor localization of UAVs will enable us to remove the dependency on expensive MoCAP systems.

ACKNOWLEDGMENT

**This work is a part of the project 45KI02B021 "Silicon Economy Logistics Ecosystem " funded by the German Federal Ministry of Transport and Digital Infrastructure.**

REFERENCES

[1] B. Öztuna, "Logistics 4.0 and Technologic Applications. In Logistics 4.0 and Future of Supply Chains," in *Springer*, Singapore, 2022.

[2] "EU Science Hub," [Online]. Available: https://joint-research-centre.ec.europa.eu/scientific-activities-z/transport-sector-economic-analysis_en. [Accessed 15 April 2022].

[3] C. G. C. D. C. I. O. A. P. P. S. G. .. &. T. M. Amza, "Guidelines on Industry 4.0 and Drone Entrepreneurship for VET students," Danmar Computers, Drone technology training to boost EU entrepreneurship and Industry, 2018.

[4] L. M. O. &. N. T. (. Wawrla, "Applications of drones in warehouse operations.," in *ETH Zurich, D-MTEC*, 2018.

[5] D. B. F. T. A. &. D. F. M. Cristiani, "Inventory management through mini-drones: Architecture and proof-of-concept implementation," in *IEEE*, 2020.

[6] C. S. &. V. L. P. ang, "The strategic role of logistics in the industry 4.0 era. Transportation Research Part," in *Logistics and Transportation*, 2019.

[7] R. E. a. S. C. C. Miles, "Organization theory and supply chain management: An evolving research perspective," *Journal of operations management,* vol. 25, no. 2, pp. 459--463, 2007.

[8] V. K. S. D. a. J. F. Markus Lieret, "Automated in-house transportation of small load carriers with autonomous Unmanned Aerial Vehicles,," in *2019 IEEE 15th International Conference on Automation Science and Engineering (CASE)*, IEEE, 2019, pp. 1010-1015.

[9] O. a. N. T. Maghazei, "Drones in manufacturing: Exploring opportunities for research and practice," *Journal of Manufacturing Technology Management,* vol. 31, pp. 1237-1259, 2019.

[10] D. a. A. J. a. P. N. Mourtzis, "UAVs for Industrial Applications: Identifying Challenges and Opportunities from the Implementation Point of View," *Procedia Manufacturing,* vol. 55, pp. 183--190, 2021.

[11] A. A. T. S. E. &. Y. N. Gupta, "Advances of UAVs toward Future Transportation: The State-of-the-Art, Challenges, and Opportunities. Future Transportation," in *MDPI*, 2021.

[12] A. S. B. a. M. S.-F. Daniel KD Villa, "A survey on load transportation using multirotor UAVs," *Journal of Intelligent & Robotic Systems,* vol. 98, no. 2, pp. 267-296, 2020.

[13] S. K. N. E.-A. M. M. H. Reem Alshanbari, 2019 IEEE National Aerospace and Electronics Conference (NAECON), AI Powered Unmanned Aerial Vehicle for Payload Transport Application, IEEE, 2019.

[14] M. M. S. B. K. Marius Boshoff, "Use of autonomous UAVs for material supply in terminal strip assembly," *TechRxiv,* 2022.

[15] L. A. a. A.-R. M. a. T.-T. L. M. M{\'a}rquez-Vega, "Multi-objective optimization of a quadrotor flock performing target zone search," *Swarm and Evolutionary Computation,* vol. 60, p. 100733, 2021.

[16] J. a. E. C. D. a. W. S. A. a. G. S. A. Kim, "Cooperative Sensor-based Selective Graph Exploration Strategy for a Team of Quadrotors," *Journal of Intelligent \& Robotic Systems,* vol. 103, pp. 1--14, 2021.

[17] W. H. G. S. S. a. N. A. James A Preiss, "Crazyswarm: A large nano-quadcopter swarm," in *2017 IEEE International Conference on Robotics and Automation (ICRA)*, IEEE, 2017, pp. 3299-3304.

[18] V. C.-P. J. S.-B. J.́. L. S. Marc Morenza-Cinos, "Development of an RFID inventory robot (AdvanRobot)," in *Robot Operating System (ROS)*, Springer, 2017, pp. 387-417.

[19] "F.-I. für Materialfluss und Logistik IML Dortmund," *Logistik entdecken, Magazin des Fraunhofer-Instituts für Materialfluss und Logistik IML Dortmund,* p. 10, 2019.

[20] C. W. Reynolds, "Flocks, herds and schools: A distributed behavioral model," in *Proceedings of the 14th annual conference on Computer graphics and interactive techniques*, 1987.

[21] B. Crowther, "Flocking of autonomous unmanned air vehicles," *The Aeronautical Journal,* vol. 107, no. 1069, pp. 99-109, 2003.

[22] A. S. o. L. T. U. M. UAVs, "A survey on load transportation using multirotor UAVs.," in *Journal of Intelligent & Robotic Systems,*, 2020.

[23] J. P. Š. P. &. F. M. Škrinjar, "Application of unmanned aerial vehicles in logistic processes.," Cham, 2018.

[24] A. R. M. &. t. H. M. Murrenhoff, "Steuerungskonzept für virtualisierte und lernfähige Materialflusssysteme," *Logistics Journal,* 2019.

[25] M. C.-P. V. S.-B. J. S. J. L. G. R. &. P. R. Morenza-Cinos, " Development of an RFID inventory robot (AdvanRobot).," in *Springer*, 2017.

[26] S. F. X. G. Y. L. C. &. Y. J. Liu, "Theory and Realization of Secondary Task Assignment for Multi-UAV Pickup Based on Green Scheduling," in *Complexity*, 2021.

[27] E. G. M. J. &. F. I. M. T. Companik, "Feasibility of warehouse drone adoption and implementation.," *Journal of Transportation Management,* 2018.

[28] T. &. K. K. Benarbia, "A Literature Review of Drone-Based Package Delivery Logistics Systems and Their Implementation Feasibility. Sustainability," in *MDPI*, 2021.

[29] É. B. A. K. T. G. K. T. M. D. &. G. L. Beke, "he role of drones in linking industry 4.0 and ITS Ecosystems," in *IEEE 18th International Symposium on Computational Intelligence and Informatics* , 2018.